\documentclass[10pt,twocolumn,letterpaper]{article}

\usepackage{cvpr}
\usepackage{times}
\usepackage{soul}
\usepackage{url}
\usepackage[utf8]{inputenc}
\usepackage[small]{caption}
\usepackage{graphicx}
\usepackage{amsmath}
\usepackage{booktabs}
\usepackage{algorithm}
\usepackage{algorithmic}
\usepackage{epsfig}
\usepackage{amssymb}
\usepackage{color}
\usepackage{multirow}
\usepackage{subcaption}
\usepackage{tabularx}
\usepackage{flushend}

\cvprfinalcopy 

\def\cvprPaperID{216} 

\ifcvprfinal\fi
\begin{document}

\title{Cross-Resolution Face Recognition via Prior-Aided Face Hallucination and Residual Knowledge Distillation}

\author{\normalsize{Hanyang~Kong$^{1}$, Jian~Zhao$^{2,3,4}$\thanks{Jian Zhao is the corresponding author. Homepage: \protect\url{https://zhaoj9014. github.io}. This work was done during Jian Zhao served as chief technical consultant at Pensees Pte Ltd, Singapore.}, Xiaoguang~Tu$^{2,4,5}$\thanks{Xiaoguang Tu was an intern at Pensees Pte Ltd, Singapore during this work.}, Junliang~Xing$^{6}$, Shengmei Shen$^{4}$, Jiashi~Feng$^{2}$} \\
    \small{$^{1}$Xi'an Jiaotong University,$^{2}$National University of Singapore,$^{3}$National University of Defense Technology}\\
	\small{$^{4}$Pensees Pte Ltd,$^{5}$University of Electronic Science and Technology of China} \\
		\small{$^{6}$National Laboratory of Pattern Recognition, Institute of Automation, Chinese Academy of Sciences} \\
	{\small hyokong@stu.xjtu.edu.cn, zhaojian90@u.nus.edu, xguangtu@outlook.com} \\ {\small jlxing@nlpr.ia.ac.cn, jane.shen@pensees.ai, elefjia@nus.edu.sg}}

\maketitle

\begin{abstract}
Recent deep learning based face recognition methods have achieved great performance, but it still remains challenging to recognize very low-resolution query face like 28$\times $28 pixels when CCTV camera is far from the captured subject. Such face with very low-resolution is totally out of detail information of the face identity compared to normal resolution in a gallery and hard to find corresponding faces therein. To this end, we propose a \textbf{R}esolution \textbf{I}nvariant \textbf{M}odel (RIM) for addressing such cross-resolution face recognition problems, with three distinct novelties. First, RIM is a novel and unified deep architecture, containing a \textbf{F}ace \textbf{H}allucination sub-\textbf{N}et (FHN) and a \textbf{H}eterogeneous \textbf{R}ecognition sub-\textbf{N}et (HRN), which are jointly learned end to end. Second, FHN is a well-designed tri-path \textbf{G}enerative \textbf{A}dversarial \textbf{N}etwork (GAN) which simultaneously perceives facial structure and geometry prior information, \emph{i.e.} landmark heatmaps and parsing maps, incorporated with an unsupervised cross-domain adversarial training strategy to super-resolve very low-resolution query image to its 8$\times $ larger ones without requiring them to be well aligned. Third, HRN is a generic \textbf{C}onvolutional \textbf{N}eural \textbf{N}etwork (CNN) for heterogeneous face recognition with our proposed residual knowledge distillation strategy for learning discriminative yet generalized feature representation. Quantitative and qualitative experiments on several benchmarks demonstrate the superiority of the proposed model over the state-of-the-arts. Codes and models are available at
\protect\url{https://github.com/HyoKong/Cross-Resolution-Face-Recognition}.
\end{abstract}


\section{Introduction}

In recent years, face recognition based on various deep learning architectures have acquired tremendous results under some challenging scenarios such as variations of illumination \cite{mudunuri2015low}, pose \cite{zhao2018towards} and age \cite{zhao2018look}. However, different resolution, especially with very large resolution gap between query and gallery images, is also a problem which needs to be solved in real-world application. To be specific, query images are always low-resolution because of the limitation of camera performance or far shooting distance between camera and subject of interest, while pre-enrolled face images in database are all high-resolution. So how to match very \textbf{l}ow-\textbf{r}esolution (LR) queries with \textbf{h}igh-\textbf{r}esolution (HR) gallery images would be a problem worth considering.

In this work, we focus on the problem of cross-resolution face recognition. Most of the existing solutions can be separated into two categories. One is to reconstruct HR query images from LR ones before recognition \cite{huang2017wavelet,wu2016deep,zhang2018super}, which is called hallucination method. Although face hallucination can generate missing facial details, it is not directly optimized for recognition but reconstruction, thus the hallucinated faces may not be optimal for recognition. The other category is to transform LR query images and corresponding HR gallery images into a common domain invariant subspace \cite{ge2019low,yang2017discriminative,song2018adversarial}, which can take full advantage of identity information to learn a discriminative representation. Nevertheless, naively learning from LR query images especially with very low resolution can be problematic due to the absence of facial details, which would cause the learned face recognition model fail to extract discriminative features and have an unideal generalization ability.

\begin{figure*}[!htbp]
	\begin{center}
		\includegraphics[width=0.8\linewidth]{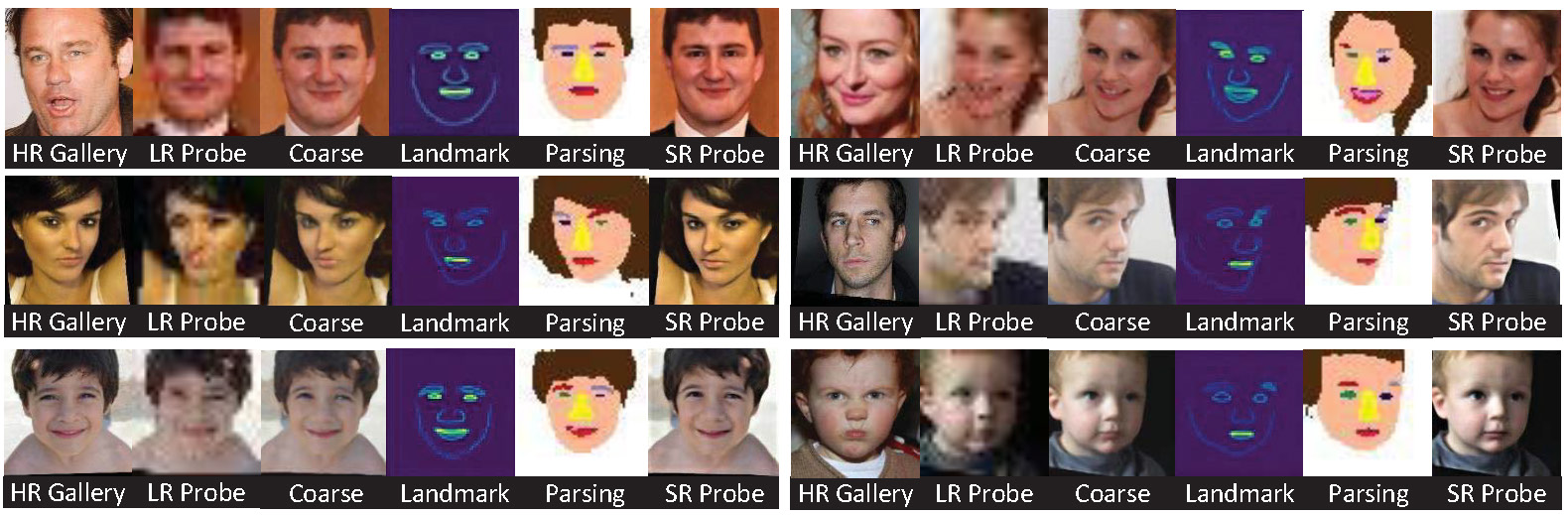}
	\end{center}
	\vspace{-4mm}
	\caption{\small Cross-resolution face recognition in the wild. Our proposed RIM can learn resolution-invariant face  representation and recover super-resolution faces efficiently with the aid of easy-to-collect prior estimation (\textit{i.e.} landmark heatmaps and parsing maps), and unsupervised domain adversarial training strategy. Best viewed in color.}
	\label{motivation}
\end{figure*}
Considering the advantages and limitations of the above methods, we propose a unified deep architecture, named \textbf{R}esolution \textbf{I}nvariant \textbf{M}odel (RIM), to super-resolve a very low-resolution probe to its 8$\times $ larger one and learn domain invariant feature representation between images with different resolution simultaneously. RIM takes a LR probe image and corresponding HR gallery one as a paired input. It outputs a reconstructive \textbf{s}uper-\textbf{r}esolution (SR) probe with the help of prior estimation, \emph{i.e.} landmark heatmap/parsing map, and meanwhile preserves discriminative representation across different identities, which offers a strong robustness to resolution variations, as illustrated in Fig. \ref{motivation}.

In particular, RIM consists of a \textbf{F}ace \textbf{H}allucination sub-\textbf{N}et (FHN) and a \textbf{H}eterogeneous \textbf{R}ecognition sub-\textbf{N}et (HRN). FHN employs a tri-path prior-aided generator with the aid of facial geometry estimation to better reconstruct the lost high-frequency information. These three pathways focus on the inference of global structure, landmark heatmap and parsing map, respectively. After that, the concatenated feature maps of global structure and prior estimation are fed into the mix-adversarial discriminator to finally reconstruct LR probes to SR one, while simultaneously maximizing the \textbf{M}ulti-\textbf{K}ernel \textbf{M}aximum \textbf{D}iscrepancy (MK-MMD) in an adversarial manner to learn feature representation invariant to the covariate shift between domains.
HRN is utilized for face verification via residual knowledge distillation between HR gallery image and corresponding SR probes recovered from LR one by FHN. In contrast to the vanilla knowledge distillation methods \cite{ashok2017n2n,Belagiannis2018AdversarialNC}, we further introduced a teacher assistant network to compensate residual error between the transferred knowledge (feature maps) of student and teacher network. With this strategy, the final output feature map is more similar to that of teacher network, hence guaranteeing excellent recognition accuracy for cross-resolution face recognition.

Our contributions are summarized as follows:

\begin{itemize}
\item We propose a unified deep architecture to achieve super-resolution face reconstruction and cross-resolution face recognition jointly.
\item We design a novel face hallucination network that can super-resolve LR images to SR ones with the aid of prior knowledge estimation and cross-domain adversarial learning strategy.
\item We develop an effective and novel training strategy, \emph{i.e.} residual knowledge distillation, for the recognition network, which can efficiently transfer knowledge between images with different resolution and generate powerful face representation.
\end{itemize}

Based on the above technical contributions, we have presented a high performance cross-resolution face recognition system which obtains competing performances over many state-of-the-art methods.

\section{Related Work}

\paragraph{Generic Face Recognition} Face recognition via deep learning has achieved a series of breakthrough in these years  \cite{sun2014deep,sun2014deep1,deng2018arcface,wen2016discriminative}. For instance, DeepID \cite{sun2014deep} and DeepID2 \cite{sun2014deep1} can be effectively learned by challenging multi-class identification and verification jointly, which achieve excellent recognition performance. Wen \emph{et al.} \cite{wen2016discriminative} propose a center loss to further enhance the capacity of discriminative feature learning. Deng \emph{et al.} \cite{deng2018arcface} introduce an additive angular margin loss to obtain highly discriminative features for face recognition. Generally speaking, deep learning models have achieved outstanding results on face recognition. However, these methods hardly perform satisfactory on cross-resolution face recognition because of the absence of facial details in very low-resolution face images.

\paragraph{Face Hallucination}
Face hallucination aims to reconstruct a HR image from a LR input, which is a domain-specific problem. Recently, various face hallucination methods based on deep convolutional neural networks have obtained the state-of-the-art performance. For example, Jiwon Kim \emph{et al.} \cite{kim2016accurate} utilize a very deep convolutional network by cascading many small filters to extract contextual information over HR images. Lai \emph{et al.} \cite{lai2017deep} propose a Laplacian pyramid super resolution network to reconstruct HR images based on cascade of convolutional neural network and residual error between upsampled feature maps and the ground truth HR images at the respective level.

Chen \emph{et al} \cite{chen2018fsrnet} recover LR images with the help of geometry prior estimation, \emph{i.e.} facial landmark heatmaps and parsing segmentation information.
Although the above methods can recover LR images to make up for missing facial details, their optimization objectives are not tailored for recognizing faces but reconstructing LR images, which will consume computing resources and affect the recognition efficiency.

\paragraph{Knowledge Distillation}
Knowledge distillation \cite{hinton2015distilling} is one of the most efficient methods for model compression and knowledge transfer, which aims at training a smaller network to mimic a more complex teacher network. Most knowledge methods typically apply one teacher network to supervise one student network. For instance, Ashok \emph{et al} \cite{ashok2017n2n} use reinforcement learning to prune student network under the guidance of teacher network. Belagiannis \emph{et al} \cite{Belagiannis2018AdversarialNC} apply adversarial strategy to knowledge distillation. They further utilize a discriminator to measure whether student model and teacher model are close enough. It is worth considering, however, that there is still a certain discrepancy between the learning capacity of teacher network and student network. Inspired by residual representation \cite{he2016deep}, we adopt an additional teacher assistant network to learn the representation gap between teacher network and student network at different-level features.

\section{Resolution-Invariant Face Recognition Model}

\begin{figure*}[!htbp]
	\begin{center}
		\includegraphics[width=0.85\linewidth]{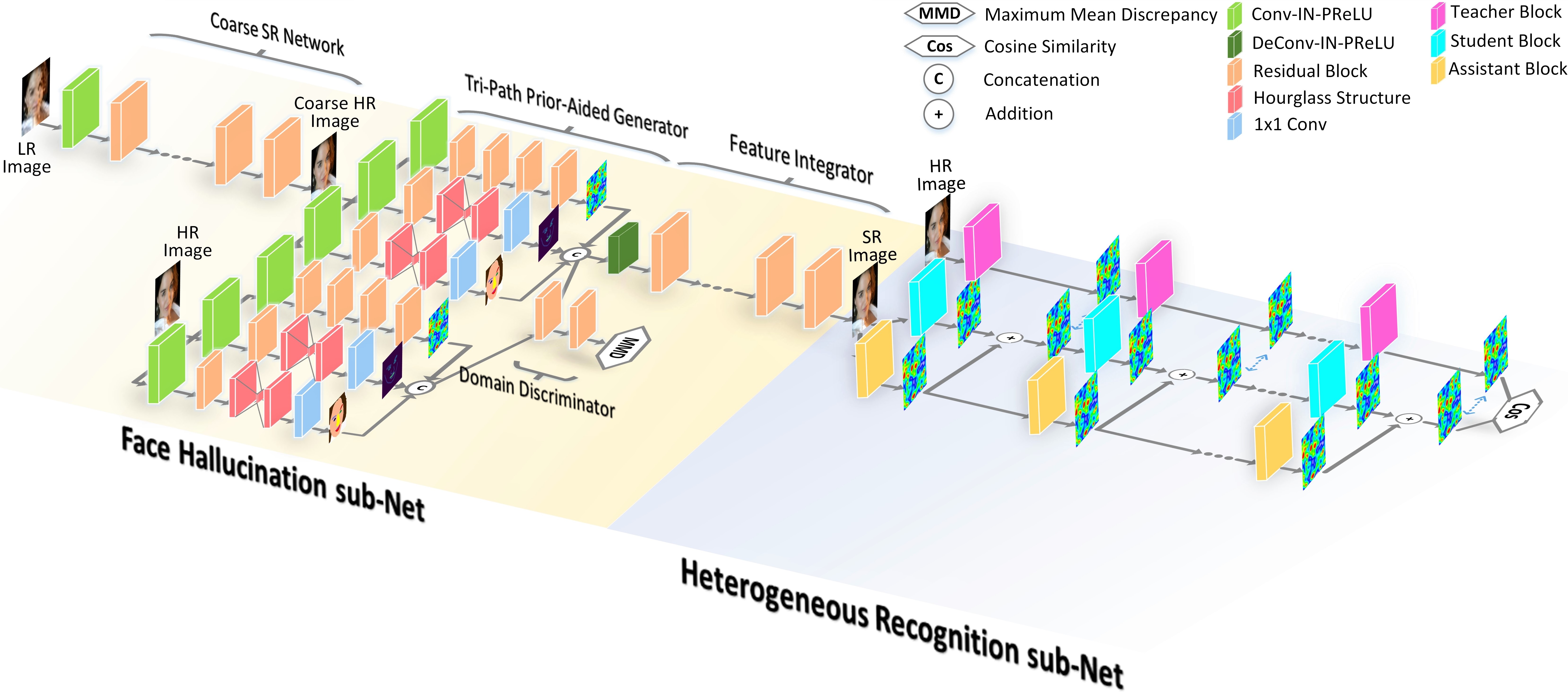}
	\end{center}
	\caption{\small \textbf{R}esolution \textbf{I}nvariant \textbf{M}odel (RIM) for face recognition in the wild. The RIM contains a \textbf{F}ace \textbf{H}allucination sub-\textbf{N}et (FHN) and a \textbf{H}eterogeneous \textbf{R}ecognition sub-\textbf{N}et (HRN) that jointly learn end-to-end. Best viewed in color.}
	\label{network architecture}
\end{figure*}

\subsection{Face Hallucination Sub-Net}

As illustrated in Fig. \ref{network architecture}, \textbf{F}ace \textbf{H}allucination sub-\textbf{N}et (FHN) consists of a prior-aided tri-path generator and a mix-adversarial discriminator. The prior-aided tri-path generator is utilized to extract feature maps of facial structure and prior knowledge, \emph{i.e.} landmark heatmaps and parsing maps, and integrate them together as the concatenated feature maps.
We further apply a mix-adversarial discriminator to reconstruct SR images from the concatenated feature maps and reduce domain gap between images with different resolution in an adversarial manner. With the iterative adversarial training strategy, FHN can learn feature representations invariant to domain shift between images with different resolution. We now present FHN in details.

Different from face recognition over a single image, the task of set-based face recognition aims to accept or reject the claimed identity of a subject represented by a face media set containing both images and videos. Performance is assessed using two measures: percentage of false accepts and that of false rejects. A good model should optimize both metrics simultaneously. MPNet is designed to nonlinearly map the raw sets of faces to multiple prototypes in a low dimensional space such that the distance between these prototypes is small if the sets belong to the same subject, and large otherwise. The similarity metric learning is achieved by training MPNet with two identical CNN branches that share weights. MPNet handles inputs in a pair-wise, set-to-set way so that it explicitly organizes the face media in a way favorable to set-based face recognition.

\subsubsection{Prior-Aided Tri-Path Generator}

Due to the deficiency of facial details and high-frequency information, face recognition using LR face image as the input query will badly affect recognition accuracy. The generic scheme to solve this problem is to super-resolve LR probes to SR counterparts before recognition. Different from general approaches that typically adopt one single CNN model for reconstruction, which cannot capture facial spacial information adequately, we further apply geometry prior knowledge, \emph{i.e.} facial landmark heatmaps and parsing maps, to aid the process of LR face reconstruction, as inspired by \cite{chen2018fsrnet}.

To be specific, the prior-aided tri-path generator $G_{\theta _{G}}$ consists of a coarse SR network $G_{\theta _{c}}$ and a tri-path generator: global feature extraction path $G_{\theta _{f}}$, landmark path $G_{\theta _{l}}$ and parsing map path $G_{\theta _{p}}$. Given a pair of face images with different resolution, we first apply $G_{\theta _{c}}$ to roughly super-resolve the LR probe to a coarse HR one which is benefit for subsequent feature extraction and prior estimation. Then the coarse HR probe and corresponding real HR image are fed to the tri-path generator simultaneously, to extract global image feature maps and estimate prior information, \emph{i.e.} landmark heatmaps and parsing maps, respectively. Finally, these output feature maps are concatenated as the output feature maps of $G_{\theta _{G}}$.

Formally, we denote the real HR face, corresponding LR and super-resolved one as $I_{hr}$, $I_{lr}$ and $I_{sr}$, respectively. After coarsely super-resolving $I_{lr}$ to a coarse HR image $I_{c}$, $I_{c}$ is fed to $G_{\theta _{G}}$ to extract the concatenated feature maps. There are two key requirements to guarantee the performance of $G_{\theta _{G}}$: 1) $G_{\theta _{G}}$ need to learn global facial feature maps and estimate geometry priors, so as to recover the SR face image better which visually resemble a real one and maintain the original identity information and textures. 2) The data distribution between source domain (with HR images) and target domain (with LR images) should be consistent, so as to make reconstructed image more similar to the real one.

To this end, we propose to learn the parameter ${\theta _{G}}$ of $G_{\theta _{G}}$ by minimizing the following combined loss:
\begin{equation}
\small
    \mathcal{L}_{\theta _{G}}=\mathcal{L}_{\mathrm{domain}}+\lambda _{0}\mathcal{L}_{\mathrm{pixel}}+\lambda _{1}\mathcal{L}_{\mathrm{landmark}}+\lambda _{2}\mathcal{L}_{\mathrm{parsing}},
\end{equation}
where $\mathcal{L}_{\mathrm{domain}}$ is the unsupervised domain adversarial loss for domain adaption between images with different resolution, $\mathcal{L}_{\mathrm{pixel}}$ is the pixel-wise Euclidean loss for constraining the visual performance and content consistency, $\mathcal{L}_{\mathrm{landmark}}$ is landmark heatmap loss for enforcing structural consistency, $\mathcal{L}_{\mathrm{parsing}}$ is the cross-entropy loss for facial parsing segmentation, and $\{\lambda_k\}_{k=_0^2}$ are weighting parameters among different losses.

In order to enhance the visual quality of super-resolved images, we apply $\mathcal{L}_{pixel}$ to constrain the consistency between the hallucinated faces and corresponding HR ground truths by penalizing the pixel-wise Euclidean distance between them:
\begin{equation}
\small
    \mathcal{L}_{\mathrm{pixel}}=\sum_{i,j}( I_{sr}^{(i,j)}-I_{hr}^{(i,j)} )^{2},
    \label{pixel}
\end{equation}
where $I_{sr}^{(i,j)}$ and $I_{hr}^{(i,j)}$ are super-resolved face and the corresponding HR ground truth at pixel $(i,j)$.

The landmark heatmap loss $\mathcal{L}_{\mathrm{landmark}}$ is introduced for landmark path $G_{\theta _{l}}$ to enforce the facial structural consistency of hallucinated face. Note that rather than training the landmark path to regress $x$ and $y$ landmark coordinates, the landmark is represented by a set of heatmaps. Each landmark point is represented by an output channel which contains a 2D Gaussian distribution centered at corresponding landmark point. The landmark path is trained to regress the 2D Gaussian map (heatmap). According to above discussion, we apply the landmark heatmap loss to enforce SR probes and corresponding HR images yielding the same heatmaps while capturing spatial context and structural part relationships, which is defined as:
\begin{equation}
\small
    \mathcal{L}_{\mathrm{landmark}}=\frac{1}{N}\sum_{n=1}^{N}\sum_{i,j}( H_{sr}^{n,(i,j)}-H_{hr}^{n,(i,j)} )^{2},
\end{equation}
where $H_{sr}^{n,(i,j)}$ represents the landmark heatmap corresponding to the $n^{th}$ landmark channel at pixel $(i,j)$, $H_{lr}^{n,(i,j)}$ represents the ground truth landmark heatmap, and $N$ denotes the number of landmark points.

Parsing segmentation loss $\mathcal{L}_{\mathrm{parsing}}$ is a pixel-wise cross-entropy loss which examines each pixel individually and compares the class prediction with the one-hot encoded vector. The parsing segmentation loss can be calculated as:
\begin{equation}
\small
    \mathcal{L}_{\mathrm{parsing}}=-\frac{1}{C}\sum_{c=1}^{C}y_{ij}log(\widetilde{y}_{ij}),
\end{equation}
where $\widetilde{y}_{ij}$ and $y_{ij}$ are the predicted value and one hot label at pixel $(i,j)$, respectively.

\subsubsection{Mix-Adversarial Discriminator}

To increase the realism of hallucination images for better face recognition, it is necessary to narrow the domain gap between images with different resolution. Hence a mix-adversarial discriminator is introduced to integrate the concatenated feature maps to SR images, while reducing the domain discrepancy between $I_{hr}$ and $I_{c}$ which is roughly super-resolved by $G_{\theta _{c}}$.

Specifically, the mix-adversarial discriminator $D_{\theta _{D}}$ is composed of two pathways: a feature integrator path $D_{\theta _{i}}$ (with learnable parameters $\theta _{i}$) and a domain discriminator path $D_{\theta _{d}}$ (with learnable parameters $\theta _{d}$). $D_{\theta _{i}}$ is applied to reconstruct SR probe from concatenated features embeded with geometry prior estimation, which is extracted from $I_{c}$. Domain discriminator and tri-path generator cooperate with each other via a domain adversarial strategy, so as to reduce domain discrepancy between images with different resolutions. To this end, we propose to learn the parameter $\theta _{i}$ by minimizing the following loss:
\begin{equation}
\small
    \mathcal{L}_{D_{\theta _{i}}}=\mathcal{L}_{\mathrm{pixel}},
\end{equation}
where $\mathcal{L}_{\mathrm{pixel}}$ is a pixel-wise $\ell_2$ distance to enforce the multi-scale content consistency between the hallucinated faces and the ground truths, which is the same as Eq.~\eqref{pixel}.

The learnable parameters $\theta _{d}$ can be learned by minimizing the following loss:
\begin{equation}
\small
    \mathcal{L}_{D_{\theta _{d}}}=-\mathcal{L}_{\mathrm{domain}},
\end{equation}
where $\mathcal{L}_{\mathrm{domain}}$ is \textbf{M}ulti-\textbf{K}ernel \textbf{M}aximum \textbf{M}ean \textbf{D}iscrepancy (MK-MMD) \cite{li2015generative}, which is a non-parametric criterion to compute the mean square distance between different data by mapping them to the \textbf{R}eproducing \textbf{K}ernel \textbf{H}ilbert \textbf{S}pace (RKHS).
Formally, $\mathcal{L}_{\mathrm{domain}}$ can be represented as:

\begin{equation}
\small
    \label{mmd}
    \mathcal{L}_{\mathrm{domain}}=\left \| \frac{1}{N_{1}}\sum_{n_{1}=1}^{N_{1}}f(x_{sr}^{n_{1}}) -  \frac{1}{N_{2}}\sum_{n_{2}=1}^{N_{2}}f(x_{hr}^{n_{2}})\right \|_{\mathcal{H}}^{2},
\end{equation}

where $x_{sr}^{n_{1}}$ and $x_{hr}^{n_{2}}$ are sampled from $X_{sr}=\left \{ x_{sr}^{n_{1}} \right \}_{n_{1}=1}^{N_{1}}$ and $X_{hr}=\left \{ x_{hr}^{n_{2}} \right \}_{n_{2}=1}^{N_{2}}$. $f$ is utilized to map data samples to RKHS. Note that in Hilbert space, the norm operation is the same thing as the inner product operation, where $\left \| f(x_{sr})-f(x_{hr}) \right \|_{\mathcal{H}}^{2}=\left \langle f(x_{sr})-f(x_{hr}) ,f(x_{sr})-f(x_{hr}) \right \rangle_{\mathcal{H}}$, so Eq. \eqref{mmd} can be rewritten by kernel tricks as
\begin{equation}
\scriptsize
\begin{aligned}
    \label{mmd2}
    &L_{\mathrm{domain}} =\frac{1}{N_{1}^{2}}\sum_{n_{1}=1}^{N_{1}}\sum_{{n_{1}}'=1}^{N_{1}}f(x_{sr}^{{n_{1}}'})^{T}f(x_{sr}^{n_{1}}) \\&-  \frac{2}{N_{1}N_{2}}\sum_{n_{1}=1}^{N_{1}}\sum_{n_{2}=1}^{N_{2}}f(x_{hr}^{n_{2}})^{T}f(x_{sr}^{n_{1}}) +  \frac{1}{N_{2}^{2}}\sum_{n_{2}=1}^{N_{2}}\sum_{{n_{2}}'=1}^{N_{2}}f(x_{hr}^{{n_{2}}'})^{T}f(x_{hr}^{n_{2}})\\
&= \frac{1}{N_{1}^{2}}\sum_{n_{1}=1}^{N_{1}}\sum_{{n_{1}}'=1}^{N_{1}}k(x_{sr}^{n_{1}},x_{sr}^{{n_{1}}'})
\\&- \frac{2}{N_{1}N_{2}}\sum_{n_{1}=1}^{N_{1}}\sum_{n_{2}=1}^{N_{2}}k(x_{sr}^{n_{1}},x_{hr}^{n_{2}})
+ \frac{1}{N_{2}^{2}}\sum_{n_{2}=1}^{N_{2}}\sum_{{n_{2}}'=1}^{N_{2}}k(x_{hr}^{n_{2}},x_{hr}^{{n_{2}}'}),
\end{aligned}
\end{equation}
where $k$ is a characteristic kernel which are combined by several convex kernels $\left \{ k_{u} \right \}$. The kernel associated with the feature maps can be defined as
\begin{equation}
\small
    \label{multi_kernel}
    \mathcal{K}\overset{\bigtriangleup }{=}\left \{ k= \sum_{u=1}^{U}\beta _{u}k_{u}:\sum_{u=1}^{U}\beta _{u}=1,\beta _{u}>0,\forall u \right \},
\end{equation}
where kernel $k_{u}$ is Gaussian kernel, defined as $k_{u}(x_{sr},x_{hr})=exp(-\frac{1}{2\sigma _{u}}\left | x_{sr}-x_{hr} \right |^{2})$, where $\sigma _{u} $ is the bandwidth parameter which is used to determine statistical efficiency of MK-MMD.

In conclusion, the domain invariant representation learning between tri-path prior-aided generator $G_{\theta _{G}}$ and domain discriminator $D_{\theta _{d}}$ can be achieved by solving the following minimax problem:
\begin{equation}
\small
    \label{minimax}
    \min \limits_{\theta_{G}}\max \limits_{\theta_{d}}{\mathcal{L}_{\mathrm{domain}}}.
\end{equation}

\subsection{Heterogeneous Recognition Sub-Net}

Considering that applying the state-of-the-art face recognition model directly would only perform well on HR query and gallery images. Re-training the model with cross-resolution images would also degrade performance because of the distribution gap between images with different resolutions. So it's crucial to incorporate the information extracted from HR images to the corresponding lower-resolution one. Hence, we utilize knowledge distillation on \textbf{H}eterogeneous \textbf{R}ecognition sub-\textbf{N}et (HRN) to train a generic feature extractor.

Vanilla knowledge distillation methods \cite{ashok2017n2n,Belagiannis2018AdversarialNC} apply a student network to learn from a large and complex network which can extract discriminative features. Considering the gap in learning capacity between teacher model and student model, there is no guarantee that student network can learn enough discriminative knowledge to reproduce relatively high performance of teacher network. So we employ an additional network, which is named teacher assistant network, to make further distillation by learning the residual error between teacher network and student network. That is to say, the assistant network is employed to help the student network fine-tune its output and transfer knowledge from teacher network more sufficiently, which is consistent to coarse-to-fine principle.

To sum up, our goal is to train a student network $N_{\theta _{S}}$ to reproduce the predictive capability of teacher network. As shown in Fig. \ref{network architecture}, we first respectively divide teacher network and student network into $K$ blocks and regard output of each block as a feature map.
Accordingly, to transfer knowledge\footnote{Feature maps are treated as knowledge in this work.} from teacher network to student network, the knowledge distillation process can be represented by the following loss
\begin{equation}
\small
    \label{student_distill}
    \mathcal{L}_{\theta _{S}}=\left \| \mathbf{f}_{T}^{K} - \mathbf{f}_{S}^{K} \right \|_{2}^{2},
\end{equation}
where $\mathbf{f}_{T}^{K}$ and $\mathbf{f}_{S}^{T}$ are respectively the output feature maps of the $K^{th}$ blocks of teacher network and student network. As shown in Eq.~\eqref{student_distill}, we can figure out that $N_{\theta _{S}}$ tries to reproduce feature maps which is same as $N_{\theta _{T}}$ to obtain powerful performance. Note that there is still a gap between feature maps extracted from $N_{\theta _{T}}$ and $N_{\theta _{S}}$, it is non-trival for $N_{\theta _{S}}$ to capture underlying feature maps with the help of $N_{\theta _{T}}$ alone. To this end, we utilize an additional network, named teacher assistant network $N_{\theta _{A}}$, to ease the differences between representation capacities of $N_{\theta _{T}}$ and $N_{\theta _{S}}$. To be specific, $N_{\theta _{A}}$ is also divided into $K$ blocks and optimized to learn residual errors between $N_{\theta _{T}}$ and $N_{\theta _{S}}$ at each corresponding blocks. Accordingly, loss function of $N_{\theta _{A}}$ can be formulated as
\begin{equation}
\small
    \label{assistant_distill}
    \mathcal{L}_{\theta _{A}}=\sum_{k=1}^{K}\left \| (\mathbf{f}_{T}^{k} - \mathbf{f}_{S}^{k}) - \mathbf{f}_{A}^{k}\right \|_{2}^{2},
\end{equation}
where $\mathbf{f}_{T}^{k}$, $\mathbf{f}_{S}^{k}$ and $\mathbf{f}_{A}^{k}$ are output feature maps of teacher network, student network and assistant network at $k^{th}$ block, respectively. After assistant and student network reaching their optima, the feature map summed up with the residual error, \emph{i.e.} $\mathbf{f}_{S} = \mathbf{f}_{S}^{K} + \mathbf{f}_{A}^{K}$, will be finally applied for inference.
\subsection{Training and Inference}

The goal of RIM is to use pairs of LR face probes and their corresponding HR one to train FHN and HRN that mutually boost and jointly accomplish cross-resolution face recognition. Each separate loss plays a role as a unique supervisor within the nested structure to converge the whole framework. The training process of RIM is an end-to-end procedure which can be optimized with various loss functions based on adversarial training strategy and back-propagation algorithm. During testing, we just feed a pair of images with different resolution as the input of RIM to get their corresponding embedding features $\mathbf{f}_{T}$ and $\mathbf{f}_{S}$. Then we calculate the cosine distance between the two embedding features and compare with the threshold to determine whether the two faces belong to the same identity.

\section{Experiments}
\label{Sec4}

\paragraph{Training Dataset}

We use Helen \cite{le2012interactive} dataset as training set. Helen dataset has 2{,}330 face images, each of which has a ground truth label of 194 landmarks and 11 parsing maps. We first perform face alignment and crop each image to $224\times 224$ by using MTCNN \cite{zhang2016joint} according to the ground truth of 5 facial landmarks and then flip each image horizontally as data augmentation.
After pre-processing, we use 2{,}800 images for model training RIM and another 1{,}100 for testing the performance of face hallucination and cross-resolution face recognition.

\paragraph{Testing Dataset}
We evaluate performance of RIM on face hallucination and cross-resolution face recognition over Helen \cite{le2012interactive}, LFW \cite{huang2008labeled}, CALFW \cite{zheng2017cross} and CPLFW \cite{sengupta2016frontal} datasets, respectively. For face hallucination, we use the Helen test set and part of LFW dataset. LFW dataset contains 13{,}233 face images with 5{,}749 identities. We randomly choose 1{,}000 aligned faces to evaluate face hallucination performance of RIM.
For cross-resolution face recognition, the input of RIM is a cross-resolution image pair and the resolution of the image pair is $224\times 224$ and $28\times 28$, respectively. We select 500 image pairs of the same identity and 500 image pairs of different identities from the Helen test set as the face recognition test set. We random choose 1500 pairs of faces of the same identity and 1500 pairs of faces of different identities as another test set of face hallucination and LR face recognition. Furthermore, we also apply CALFW \cite{zheng2017cross} and CPLFW \cite{sengupta2016frontal} datasets to evaluate the performance of cross-resolution face recognition.

\paragraph{Implementation details}
We first enlarge LR images from $28\times 28$ to $224\times 224$ by bicubic interpolation as the input of our model. RMSprop solver is applied with an initial learning rate $1\times 10^{-3}$ for all sub-net.  We apply a pre-trained ResNet50-IR \cite{deng2018arcface} face recognition model \cite{facerepo} as the teacher model and choose ResNet34 to serve as student network and assistant network in HRN. According to ResNet sturcture, 4 blocks is divided for each network. To compute residual error between teacher network and student network, outputs of corresponding blocks from different network must have the same size, especially channel dimension. We implement all the experiments by extending the publicly available PyTorch framework \cite{pytorch} on a single NVIDIA TITAN Xp GPU with 12 G memory.

\subsection{Ablation Study}

To clarify the role of each component in our model structure, we combine different sub-nets together and train them respectively for face recognition on LFW dataset. Various combinations and corresponding results are reported in Tab. \ref{Ablation Study}.

We can draw the following conclusions. First, experiment (a), (b) and (c) utilize HR image as input directly to train the teacher network, student network and distill knowledge from the teacher network to student network. The results show that we can't extract discriminative feature representation from LR images directly. In experiment (d), (e) and (f), FHN is applied to reconstruct LR images to SR images before face recognition. Comparative results of (d) and (g) show that using SR image reconstructed by face hallucination and HR image as the input of teacher network can achieve the same recognition accuracy. (e) and (f) indicate that after knowledge distillation, the recognition accuracy of the student network decreases to a certain extent because the complexity of the student network used for identification is much less than that of the teacher network. After introducing the teacher assistant network in HRN, the recognition accuracy is improved by $2\%$ compared with (e).

\begin{center}
\tiny
\begin{table*}[t]

  \centering
  \tiny
  \captionsetup{font={small}}
  \caption{Analysis of the role of each component. We run the experiment by combining different sub-nets as indicated by the checkmarks.}\footnotesize
  \scalebox{1.0}{
  \begin{tabular}{c|c|ccccccc}
  \multicolumn{2}{c}{\multirow{1}*{ }}& (a)& (b)&  (c)&  (d)&  (e)&  (f)&  (g)   \\
  \hline
  \multicolumn{2}{c|}{\multirow{1}*{Gallery-Query}}& HR-LR&  HR-LR&  HR-LR&  HR-LR&  HR-LR&  HR-LR&  HR-HR   \\
  \hline
  \hline
  \multicolumn{2}{c|}{\multirow{1}*{FHN}}&  &   &   &  \checkmark&  \checkmark&  \checkmark&      \\


  \multicolumn{1}{c|}{\multirow{3}*{HRN}} & T: ResNet50-IR&  \checkmark&   &  \checkmark&  \checkmark&  \checkmark&  \checkmark&  \checkmark   \\
  & S: ResNet34&   &   \checkmark&  \checkmark&   &   &  \checkmark&     \\
  & A: ResNet34&   &   &   &   &   &  \checkmark&     \\
  \multicolumn{2}{c|}{\multirow{1}*{Acc}}&  0.843&   0.687&   0.802&  0.997&  0.966&  0.988&   0.998\\
  \hline
  \end{tabular}}
  \label{Ablation Study}
\end{table*}
\end{center}

\subsection{Comparisons with the State-of-the-Arts}

\subsubsection{Evaluation on Cross-Resolution Face Recognition}

We compare our method with the state-of-the-art face recognition methods \cite{deng2018arcface,cao2018vggface2,wen2016discriminative} to evaluate the performance of cross-resolution face recognition.
ArcFace \cite{deng2018arcface} proposes an additive angular margin loss to obtain highly discriminative features using a clear geometric interpretation. VGGFace2 \cite{cao2018vggface2} utilizes ResNet50 \cite{he2016deep} to access performance on face recognition. CenterFace \cite{wen2016discriminative} proposes center loss to recognize identities which learns a center for deep features of each class and penalizes the distances between the deep features and their corresponding class centers. We respectively utilize HR image, SR image and LR images as the input of these methods and compare the recognition results with our method for cross-resolution face recognition. All results are reported in Tab. \ref{recognition_result}.
Note that \cite{deng2018arcface,cao2018vggface2,wen2016discriminative} are all pre-trained on HR images.


\renewcommand{\multirowsetup}{\centering}
\begin{table*}[!htbp]
\small
\captionsetup{font={small}}
\caption{Cross-resolution face identification evaluation on four different datasets.}\footnotesize
    \centering
    \scalebox{1.0}{
    \begin{tabular}{ccccccc}
    \hline
    &   Gallery-Query& Method&  Helen\cite{le2012interactive}&    LFW\cite{huang2008labeled}&   CALFW\cite{zheng2017cross}&    CPLFW\cite{sengupta2016frontal}\\
    \hline
    \hline
         (a)&  HR-HR&    ArcFace\cite{deng2018arcface}&  0.998&  0.998&  0.952&    0.921\\
         (b)&  HR-HR&    VGGFace2\cite{cao2018vggface2}&  -&  0.994&  -&    0.840\\
         (c)&  HR-HR&    Centerface\cite{wen2016discriminative}&  -&  0.988&  -&    0.775\\
         (d)&  HR-SR&    ArcFace\cite{deng2018arcface}&  0.998&  0.998&  0.950&    0.920\\
         (e)&  HR-SR&    VGGFace2\cite{cao2018vggface2}&  -&  0.990&  -&    0.827\\
         (f)&  HR-SR&    Centerface\cite{wen2016discriminative}&  -&  0.975&  -&    0.763\\
         (g)&  HR-LR&    ArcFace\cite{deng2018arcface}&  0.843&  0.807&  0.781&    0.797\\
             (h)&  HR-LR&    \textbf{RIM(Ours)}&  \textbf{0.987}&  \textbf{0.988}&  \textbf{0.943}&  \textbf{0.908}\\
    \hline
\end{tabular}}
\label{recognition_result}
\end{table*}
From Tab. \ref{recognition_result}, we can draw the following conclusions: (1) For almost all methods above, compared with recognition results which input pairs are both HR images, using the hallucination SR/HR images as input hardly reduce the recognition accuracy of the models. The possible explanation is that plenty facial details reconstructed from LR face images make it easier for model to extract discriminative features, which is benefical for recognition performance. (2) As shown in (g), directly using HR-LR image pairs as input of ArcFace \cite{deng2018arcface} for face recognition will dramatically decrease the recognition accuracy even though the model has powerful capacity to extract discriminative features. This indicates the importance of face hallucination for LR images in cross-resolution face recognition tasks. (3) The performance of our proposed network structure in cross-resolution face recognition is very close, and even can exceed than that of various HR-HR face recognition methods. This benefits from two aspects: For one thing, with the help of prior information and domain adaption between LR and HR images, our network can obtain high-fidelity reconstructive SR images which offer sufficient facial details. For another, instead of training a recognition network directly, we adopt residual knowledge distillation to transfer knowledge from teacher network to student network. To transfer knowledge more efficiently, an assistant network is incorporated to make up the gap between teacher network and student network.
\renewcommand{\multirowsetup}{\centering}
\begin{center}
\begin{table*}[htbp]
\small
  \centering
  \captionsetup{font={small}}
  \caption{Cross-resolution face verification on LFW \cite{huang2008labeled}.}\footnotesize
  \scalebox{1.0}{
  \begin{tabular}{p{0.9cm}<{\centering}p{3.0cm}<{\centering}p{3.1cm}<{\centering}p{1.8cm}<{\centering}p{1.8cm}<{\centering}}
  \hline
  \multicolumn{1}{c}{\multirow{2}*{ }}&\multicolumn{1}{c}{\multirow{2}{2.8cm}{Method}} &\multicolumn{1}{c}{\multirow{2}{2.6cm}{Training Data}}   & \multicolumn{2}{c}{Testing (Gallery-Query)}  \\
    \cline{4-5}
    \multicolumn{3}{c}{}&HR-HR&HR-LR\\
  \hline
  \hline
      (a)& Teacher Network& HR&  0.998&      0.843\\
      (b)& Student Network& LR&  -&     0.807\\
      (c)& Staged-CNN\cite{peng2016fine}& HR \& LR&  0.908&    0.902  \\
      (d)& Guided-CNN\cite{fu2017learning}& HR \& LR&  0.974&     0.938 \\
      (e)& \textbf{RIM(Ours)}& HR \& LR&  \textbf{0.998}&    \textbf{0.988}\\
  \hline
  \end{tabular}}
\label{cross_resolution}
\end{table*}
\end{center}
\vspace{-0.8cm}
Furthermore, we compare our proposed method with two baselines and other cross-resolution face recognition methods on LFW dataset. \cite{peng2016fine} applies the CNN model with staged-training to address this problem. They utilize a simple stage-wise training procedure that first trains the model on HR images and artificially lowers the resolution of training images to reduce the domain gap between images with different resolution. \cite{fu2017learning} utilizes parallel sub-CNN model as guide and learners for cross-resolution recognition. From Tab. \ref{cross_resolution}, we can observe that only train baseline on HR or LR images would not perform well on cross-resolution face recognition. \cite{peng2016fine} and \cite{fu2017learning} improve the recognition accuracy by transfer knowledge from HR images to LR ones. It is worth noting that, compared with other baselines or methods under the setting of HR-HR face verification,, RIM achieves comparable performance under a much more challenging setting of HR-LR, which outperforms the $2^{nd}-best$ by 0.05.

\subsubsection{Evaluation on Face Hallucination}

\begin{figure*}[!htbp]
\small
    \centering
    \captionsetup{font={small}}
    \includegraphics[scale=0.8]{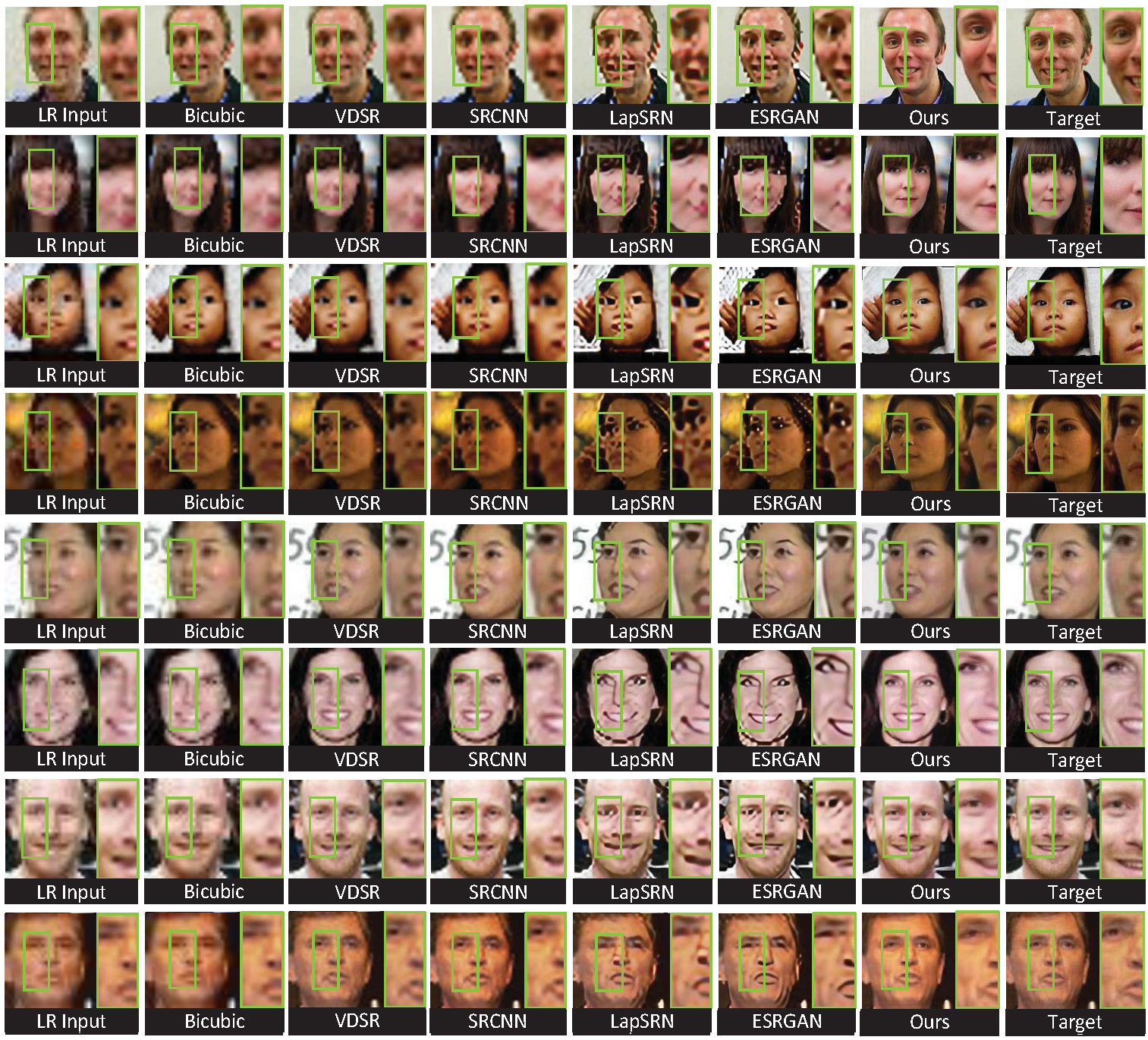}
    \caption{Qualitative comparision on Helen \cite{le2012interactive} and LFW \cite{huang2008labeled} datasets. The first row is sampled from Helen and the bottom is sampled from LFW.}
    \label{Heterogeneous}
\end{figure*}

\renewcommand{\multirowsetup}{\centering}
\begin{table*}[!htbp]
\small
\captionsetup{font={small}}
    \centering
    \caption{Quantitative evaluation on LFW \cite{huang2008labeled} dataset with PSNR/SSIM criterion.}
    \scalebox{1.0}{
    \begin{tabular}{ccccccc}
    \hline
    Method&  Bicubic&    VDSR\cite{kim2016accurate}&   SRCNN\cite{dong2015image}&   LapSRN\cite{lai2017deep}&  ESRGAN\cite{wang2018esrgan}&  \textbf{RIM(Ours)}\\
    \hline
    \hline
         PSNR&  24.965&  25.613&  25.712&  26.380&  25.607& \textbf{28.572}\\
         SSIM&  0.629&  0.730&  0.713&  0.786&  0.783&  \textbf{0.882}\\
    \hline
    \end{tabular}}
\label{criterion}
\end{table*}

\begin{figure*}[!htbp]
\small
    \centering
    \captionsetup{font={small}}
    \includegraphics[scale=0.55]{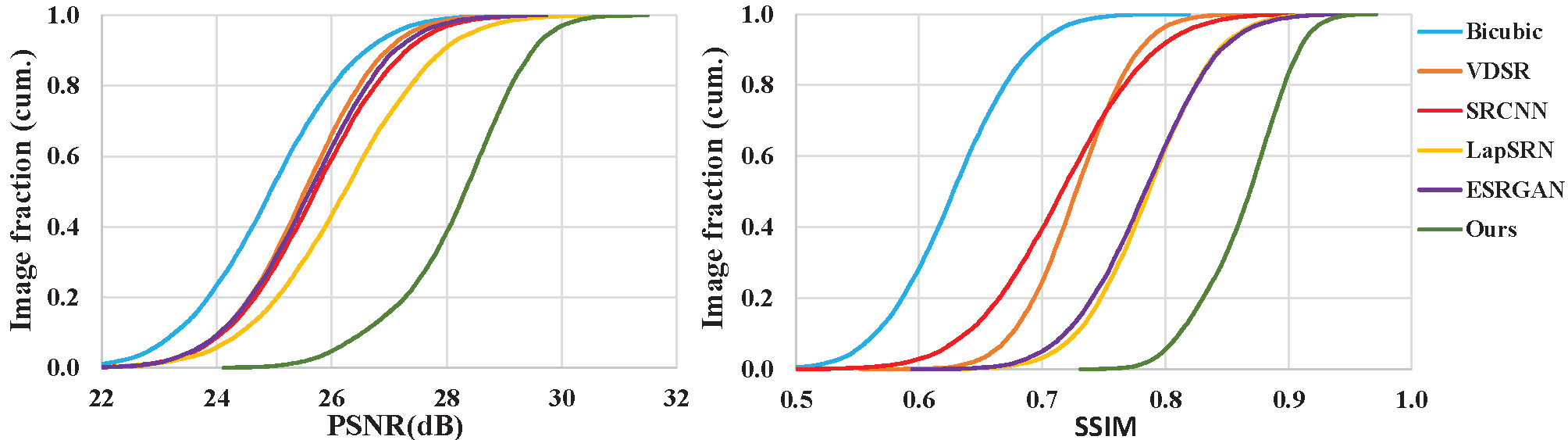}
    \caption{\textbf{C}umulative \textbf{S}core \textbf{D}istribution (CSD) scores for PSNR (left) and SSIM (right) on LFW \cite{huang2008labeled} datasets. Curves further to the right are of better quantitative performance. Best viewed in color.}
    \label{csd}
\end{figure*}

For qualitative comparison, we compare our face hallucination performance with four state-of-the-art methods: VDSR \cite{kim2016accurate}, SRCNN \cite{dong2015image}, LapSRN \cite{lai2017deep} and ESRGAN \cite{wang2018esrgan} in Fig. \ref{Heterogeneous}. For fair comparison, we train all models by their released code with the same train set. From Fig. \ref{Heterogeneous}, we can draw the following conclusion: (1) With a high magnification scale 8 $\times $, bicubic interpolation cannot provide sufficient facial details. (2) VDSR and SRCNN supplement some facial details by utilizing cascaded CNN, but fail to provide substantial texture information. (3) LapSRN and ESRGAN recover facial details and texture information better than VDSR and SRCNN, but there are still some losses of facial details and some amplified noise present in LR images. (4) Benefitting from prior information and reduction in distribution discrepancy between images with different resolutions, our method significantly outperforms other methods for face hallucination.

Furthermore, we quantitatively compute average PSNR and SSIM over LFW dataset, as reported in Tab. \ref{criterion}.
Our method outperforms the $2^{nd}-best$ by 2.192dB and 0.096 in terms of PSNR and SSIM, respectively. Furthermore, to get better insight into the performance, we present \textbf{C}umulative \textbf{S}core \textbf{D}istribution (CSD) curves for PSNR and SSIM, as illustrated in Fig. \ref{csd}. We can observe that there is an obvious gap between the quantitative results of face hallucination with different methods. VDSR and SRCNN have similar quantitative results in terms of PSNR and SSIM criterion, which are consistent with their visualization. LapSRN and ESRGAN are very close in terms of SSIM-based CSD curves, but the difference become larger with PSNR criterion. By comprehensive comparison, our proposed PIM is superior to other methods for super-resolving LR faces. Both qualitative and quantitative comparisons clearly demonstrate the superiority of our proposed RIM for face hallucination.

\section{Conclusion}

We propose a novel \textbf{R}esolution-\textbf{I}nvariant \textbf{M}odel (RIM) to address the challenging cross-resolution face recognition problem. RIM unifies a \textbf{F}ace \textbf{H}allucination sub-\textbf{N}et (FHN) and a \textbf{H}eterogeneous \textbf{R}ecognition sub-\textbf{N}et (HRN) for resolution-invariant recognition in an end-to-end deep architecture. The FHN introduce a well-designed face hallucination model with the aid of geometry prior knowledge, \emph{i.e.} facial landmark heatmaps and parsing maps, to super-resolve low-resolution query to its $8 \times$ larger one with recovered high-fidelity facial details. The HRN introduces a generic convolutional neural network with a new residual knowledge distillation strategy. Comprehensive experiments demonstrate the superiority of RIM over the state-of-the-arts.

\section*{Acknowledgement}

The work of Jian Zhao was partially supported by \textbf{C}hina \textbf{S}cholarship \textbf{C}ouncil (CSC) grant 201503170248.

The work of Junliang Xing was partially supported by the National Science Foundation of China 61672519.

The work of Jiashi Feng was partially supported by NUS IDS R-263-000-C67-646, ECRA R-263-000-C87-133 and MOE Tier-II R-263-000-D17-112.

{\small
\bibliographystyle{ieee}
\bibliography{library1}

\begin{thebibliography}{10}\itemsep=-1pt

\bibitem{ashok2017n2n}
A.~Ashok, N.~Rhinehart, F.~Beainy, and K.~M. Kitani.
\newblock N2n learning: Network to network compression via policy gradient
  reinforcement learning.
\newblock {\em arXiv preprint arXiv:1709.06030}, 2017.

\bibitem{Belagiannis2018AdversarialNC}
V.~Belagiannis, A.~Farshad, and F.~Galasso.
\newblock Adversarial network compression.
\newblock In {\em ECCVW}, 2018.

\bibitem{cao2018vggface2}
Q.~Cao, L.~Shen, W.~Xie, O.~M. Parkhi, and A.~Zisserman.
\newblock Vggface2: A dataset for recognising faces across pose and age.
\newblock In {\em FG}, pages 67--74, 2018.

\bibitem{chen2018fsrnet}
Y.~Chen, Y.~Tai, X.~Liu, C.~Shen, and J.~Yang.
\newblock Fsrnet: End-to-end learning face super-resolution with facial priors.
\newblock In {\em CVPR}, pages 2492--2501, 2018.

\bibitem{deng2018arcface}
J.~Deng, J.~Guo, N.~Xue, and S.~Zafeiriou.
\newblock Arcface: Additive angular margin loss for deep face recognition.
\newblock {\em arXiv preprint arXiv:1801.07698}, 2018.

\bibitem{dong2015image}
C.~Dong, C.~C. Loy, K.~He, and X.~Tang.
\newblock Image super-resolution using deep convolutional networks.
\newblock {\em TPAMI}, 38(2):295--307, 2015.

\bibitem{pytorch}
Facebook.
\newblock Pytorch.
\newblock \url{https://github.com/pytorch/pytorch}.
\newblock January, 2017.

\bibitem{fu2017learning}
T.-C. Fu, W.-C. Chiu, and Y.-C.~F. Wang.
\newblock Learning guided convolutional neural networks for cross-resolution
  face recognition.
\newblock In {\em MLSP}, pages 1--5, 2017.

\bibitem{ge2019low}
S.~Ge, S.~Zhao, C.~Li, and J.~Li.
\newblock Low-resolution face recognition in the wild via selective knowledge
  distillation.
\newblock {\em TIP}, 28(4):2051--2062, 2019.

\bibitem{he2016deep}
K.~He, X.~Zhang, S.~Ren, and J.~Sun.
\newblock Deep residual learning for image recognition.
\newblock In {\em CVPR}, pages 770--778, 2016.

\bibitem{hinton2015distilling}
G.~Hinton, O.~Vinyals, and J.~Dean.
\newblock Distilling the knowledge in a neural network.
\newblock {\em arXiv preprint arXiv:1503.02531}, 2015.

\bibitem{huang2008labeled}
G.~B. Huang, M.~Mattar, T.~Berg, and E.~Learned-Miller.
\newblock Labeled faces in the wild: A database forstudying face recognition in
  unconstrained environments.
\newblock In {\em ECCVW}, 2008.

\bibitem{huang2017wavelet}
H.~Huang, R.~He, Z.~Sun, and T.~Tan.
\newblock Wavelet-srnet: A wavelet-based cnn for multi-scale face super
  resolution.
\newblock In {\em ICCV}, pages 1689--1697, 2017.

\bibitem{kim2016accurate}
J.~Kim, J.~Kwon~Lee, and K.~Mu~Lee.
\newblock Accurate image super-resolution using very deep convolutional
  networks.
\newblock In {\em CVPR}, pages 1646--1654, 2016.

\bibitem{lai2017deep}
W.-S. Lai, J.-B. Huang, N.~Ahuja, and M.-H. Yang.
\newblock Deep laplacian pyramid networks for fast and accurate
  super-resolution.
\newblock In {\em CVPR}, pages 624--632, 2017.

\bibitem{le2012interactive}
V.~Le, J.~Brandt, Z.~Lin, L.~Bourdev, and T.~S. Huang.
\newblock Interactive facial feature localization.
\newblock In {\em ECCV}, pages 679--692, 2012.

\bibitem{li2015generative}
Y.~Li, K.~Swersky, and R.~Zemel.
\newblock Generative moment matching networks.
\newblock {\em arXiv preprint arXiv:1502.02761}, 2015.

\bibitem{mudunuri2015low}
S.~P. Mudunuri and S.~Biswas.
\newblock Low resolution face recognition across variations in pose and
  illumination.
\newblock {\em TPAMI}, 38(5):1034--1040, 2015.

\bibitem{peng2016fine}
X.~Peng, J.~Hoffman, X.~Y. Stella, and K.~Saenko.
\newblock Fine-to-coarse knowledge transfer for low-res image classification.
\newblock In {\em ICIP}, pages 3683--3687, 2016.

\bibitem{sengupta2016frontal}
S.~Sengupta, J.-C. Chen, C.~Castillo, V.~M. Patel, R.~Chellappa, and D.~W.
  Jacobs.
\newblock Frontal to profile face verification in the wild.
\newblock In {\em WACV}, pages 1--9, 2016.

\bibitem{song2018adversarial}
L.~Song, M.~Zhang, X.~Wu, and R.~He.
\newblock Adversarial discriminative heterogeneous face recognition.
\newblock In {\em AAAI}, 2018.

\bibitem{sun2014deep1}
Y.~Sun, Y.~Chen, X.~Wang, and X.~Tang.
\newblock Deep learning face representation by joint
  identification-verification.
\newblock In {\em NIPS}, pages 1988--1996, 2014.

\bibitem{sun2014deep}
Y.~Sun, X.~Wang, and X.~Tang.
\newblock Deep learning face representation from predicting 10,000 classes.
\newblock In {\em CVPR}, pages 1891--1898, 2014.

\bibitem{wang2018esrgan}
X.~Wang, K.~Yu, S.~Wu, J.~Gu, Y.~Liu, C.~Dong, Y.~Qiao, and C.~Change~Loy.
\newblock Esrgan: Enhanced super-resolution generative adversarial networks.
\newblock In {\em ECCV}, pages 0--0, 2018.

\bibitem{wen2016discriminative}
Y.~Wen, K.~Zhang, Z.~Li, and Y.~Qiao.
\newblock A discriminative feature learning approach for deep face recognition.
\newblock In {\em ECCV}, pages 499--515, 2016.

\bibitem{wu2016deep}
J.~Wu, S.~Ding, W.~Xu, and H.~Chao.
\newblock Deep joint face hallucination and recognition.
\newblock {\em arXiv preprint arXiv:1611.08091}, 2016.

\bibitem{yang2017discriminative}
F.~Yang, W.~Yang, R.~Gao, and Q.~Liao.
\newblock Discriminative multidimensional scaling for low-resolution face
  recognition.
\newblock {\em IEEE Signal Process Lett}, 25(3):388--392, 2017.

\bibitem{zhang2018super}
K.~Zhang, Z.~Zhang, C.-W. Cheng, W.~H. Hsu, Y.~Qiao, W.~Liu, and T.~Zhang.
\newblock Super-identity convolutional neural network for face hallucination.
\newblock In {\em ECCV}, pages 183--198, 2018.

\bibitem{zhang2016joint}
K.~Zhang, Z.~Zhang, Z.~Li, and Y.~Qiao.
\newblock Joint face detection and alignment using multitask cascaded
  convolutional networks.
\newblock {\em IEEE Signal Processing Lett}, 23(10):1499--1503, 2016.

\bibitem{facerepo}
J.~Zhao.
\newblock face.evolve: High-performance face recognition library based on
  pytorch.
\newblock \url{https://github.com/ZhaoJ9014/face.evoLVe.PyTorch}.
\newblock January, 2019.

\bibitem{zhao2018look}
J.~Zhao, Y.~Cheng, Y.~Cheng, Y.~Yang, H.~Lan, F.~Zhao, L.~Xiong, Y.~Xu, J.~Li,
  S.~Pranata, et~al.
\newblock Look across elapse: Disentangled representation learning and
  photorealistic cross-age face synthesis for age-invariant face recognition.
\newblock {\em arXiv preprint arXiv:1809.00338}, 2018.

\bibitem{zhao2018towards}
J.~Zhao, Y.~Cheng, Y.~Xu, L.~Xiong, J.~Li, F.~Zhao, K.~Jayashree, S.~Pranata,
  S.~Shen, J.~Xing, et~al.
\newblock Towards pose invariant face recognition in the wild.
\newblock In {\em CVPR}, pages 2207--2216, 2018.

\bibitem{zheng2017cross}
T.~Zheng, W.~Deng, and J.~Hu.
\newblock Cross-age lfw: A database for studying cross-age face recognition in
  unconstrained environments.
\newblock {\em arXiv preprint arXiv:1708.08197}, 2017.

\end{thebibliography}
}

\end{document}